\RequirePackage{amsmath,amsfonts}
\documentclass[graybox,openany]{svmult}
\usepackage{multirow}
\usepackage{mathptmx}

\usepackage{booktabs}
\usepackage[T1]{fontenc}

\usepackage{booktabs}
\usepackage{graphicx}
\usepackage{xcolor}
\usepackage{multirow}
\usepackage{multicol}

\begin{document}

\begingroup
\thispagestyle{empty}
\centering

\vspace*{2cm}

{\LARGE\bfseries
Explainable Reinforcement Learning for\\
Assisting Air Traffic Controllers\par}

\vspace{1.5cm}

{\large
This is the author's peer-reviewed, accepted manuscript version
(Camera-Ready) of the chapter published in the lecture notes volume below.\par}

\vspace{2cm}

\hrule
\vspace{0.8cm}

{\itshape
The final publication is available at Springer via:\\[0.3cm]
\url{https://doi.org/10.1007/978-3-031-87778-0\_14}\par
}

\vspace{0.8cm}
\hrule

\vspace{2cm}

{\small
\textbf{Full Bibliographic Citation:}\\

Mehmeti, A., Gigante, G., Venticinque, S. (2025).
\textit{Explainable Reinforcement Learning for Assisting Air Traffic Controllers}.
In: Barolli, L. (eds) \textit{Advanced Information Networking and Applications}.
AINA 2025.
Lecture Notes on Data Engineering and Communications Technologies, vol. 250.
Springer, Cham.
\url{https://doi.org/10.1007/978-3-031-87778-0\_14}

\vfill

{\footnotesize\color{gray}
Provided for personal, non-commercial, and institutional repository use in accordance with Springer Nature Self-Archiving Policy.
}
} 

\endgroup
\clearpage

\newpage
\title*{Explainable Reinforcement Learning for assisting Air Traffic Controllers}
%
%
\authorrunning{Anduel Mehmeti et Al.}
\author{Anduel Mehmeti, Gabriella Gigante, Salvatore Venticinque}
\institute
 {   Anduel Mehmeti \at  Department of Engineering, University of Campania "Luigi Vanvitelli", Italy  \email{anduel.mehmeti@unicampania.it}
         \and
    Gabriella Gigante \at CIRA, Italy \email{ggigante@cira.it}  \and
        Salvatore Venticinque \at Department of Engineering, University of Campania "Luigi Vanvitelli", Italy          \email{salvatore.venticinque@unicampania.it}
}
\maketitle
\abstract{
To effectively integrate AI into high-stakes, critical environments such as healthcare, autonomous driving, and aviation—and to advance toward higher levels of automation and seamless human-AI collaboration—building trust in AI-driven solutions is essential. Trust, in turn, is closely linked to the explainability of AI systems. The rapid advancements in AI across various domains have underscored the challenges of establishing trust, raising increasing interest in AI explainability even more when applied to deep learning. In this context, the present work aims to explore the application of explainability techniques to Reinforcement Learning (RL) algorithms, specifically within the safety-critical domain of Air Traffic Control (ATC). Using a simplified ATC environment as an initial testbed, an intelligent agent is trained with a reinforcement learning algorithm to make decisions on alternative flight routes that avoid no-fly zones. As a preliminary explainability approach, a saliency map is employed, providing insights into the input features that most significantly influence the agent’s decision-making process.
}


%
\section{Introduction}
\label{sect:intro}
To effectively integrate AI into high-stakes, critical environments such as healthcare, autonomous driving, and aviation—and to advance toward higher levels of automation and seamless human-AI collaboration—building trust in AI-driven solutions is essential. Trust, in turn, is closely linked to the explainability of AI systems (XAI). The rapid advancements in AI across various domains have underscored the challenges of establishing trust, raising increasing interest in AI explainability even more when applied to deep learning~\cite{Preece2018}. Furthermore, The EU AI Act (Artificial Intelligence Act), which is the European Union's legislative proposal aimed at regulating artificial intelligence, proposed in April 2021, categorizes AI systems based on their risk levels (e.g., high-risk, limited risk, and minimal risk) and introduces specific requirements for each category, including, in article 14,  provisions for transparency, accountability, and explainability. The regulatory bodies in different domains are addressing as an element of trustworthiness explainability.
 Explainability refers to the degree to which the internal workings and decisions of an AI system can be understood by humans. It is the ability to provide a clear explanation for how an AI system arrives at a particular decision or prediction. In complex models like reinforcement learning (RL), explainability aims to demystify the decision-making process of an agent, ensuring that users, stakeholders, and regulators understand its reasoning
 Explainability regards different stakeholder. In the aviation domain, the European aviation authority, EASA, has addressed three different types of explainability considering the different possible end-users, the dev-op explainability for the AI solution developers, the ops explainability for the AI solution end-users, and the post-op explainability for AI governance and maintenance in the organizations ~\cite{EASAconceptpaper} .  
Very Recent researches have emphasized the relevance of explainability in terms of opportunities since the related market is expected to grow of 20.9\% in the next five years   and of research challenges among them is notably considered XAI for RL techniques~\cite{saeed2023explainable}.
Air traffic management due to the growing demand has amplified the need of autonomous solutions capable of handling critical tasks strengthening the needs of supporting solutions in critical tasks. Reinforcement Learning (RL), particularly Deep Q-Network (DQN), seems to be promising in the domain ~\cite{nebula2023digital}.  However, the lack of transparency in RL models limits their applicability in safety-critical environments such as Air Traffic Control (ATC). This research focuses on developing a decision-support system that not only ensures safe and efficient en-route navigation, but also incorporates an explainability layer to improve the system's interpretability and trustworthiness for human operators.
A simplified problem in ATM domain serves as the initial testbed for this study. The environment features a 2D grid, where an intelligent agent controls a single aircraft tasked with navigating to a target location while avoiding a no-fly zone. The agent's actions involve adjusting speed and heading angle to achieve efficient and safe en-route navigation. A reward structure incentives movement toward the target, penalizing no-fly zone violations and inefficient trajectories.
The decision-making process is powered by a DQN model with separate outputs for speed and angle adjustments. The system incorporates saliency maps as a preliminary explainability technique, offering insights into the input features (position, speed, heading) most influential in the agent's decisions. These visualizations provide stakeholders with an understanding of the agent's priorities, thereby enhancing trust in its actions.
Initial results from this simplified environment demonstrate the agent's ability to perform en-route navigation toward the target while avoiding restricted zones (no-fly zones). However, ongoing work aims to refine the model's performance and extend the explainability framework. The planned enhancements include additional layers for reward decomposition, policy visualization, and counterfactual explanations to further improve interpretability.
This research underscores the potential for integrating explainability techniques into RL systems, particularly in safety-critical applications such as ATC. By bridging the gap between automation and human interpretability, the system aims to align with regulatory standards while fostering trust and collaboration between human operators and AI agents. Future work will focus on expanding the system's capabilities and validating its performance in more complex and realistic air traffic scenarios.

\section{Related Work}
\label{sect:related}

The rapid growth in global air traffic has significantly increased the complexity of air traffic management (ATM), presenting challenges in ensuring safety and efficiency. Air Traffic Control (ATC) systems are essential for ensuring smooth operations within controlled airspace. Traditionally, ATC relies heavily on human controllers to perform safety critical tasks. Although effective, these methods are struggling to meet the increasing demands of modern aviation. Reinforcement Learning (RL), a branch of machine learning, has emerged as a promising solution for automating task where few data are available and were the problem of interest involves a sequence of decisions or actions, influencing future states in dynamic environments by enabling intelligent decision-making. However, the "black-box" nature of RL models limits their interpretability, posing significant challenges for trust, adoption, and regulatory compliance in safety-critical domains like ATC.

Explainability is crucial for integrating RL into ATC systems~\cite{wang2025enhancing}. The research community has developed a very broad range of different methods and approaches and comprehensive overviews provide a clear picture of the development across the last 5 years~\cite{degas2022survey}. Several methods have been proposed to enhance the explainability of RL systems. Techniques like SHapley Additive exPlanations (SHAP)~\cite{Heuillet2021} and Local Interpretable Model-Agnostic Explanations (LIME)~\cite{Heuillet2021} assess the influence of input features (e.g., aircraft speed, altitude, position) on the model's outputs. These methods offer insights into why specific actions are recommended, aiding Air Traffic Controllers (ATCOs) in understanding how to separate the aircraft from the no-fly zone.
Using simpler, inherently interpretable models such as decision trees or linear models provides straightforward, rule-based explanations. Although these models may lack the predictive power of deep neural networks, they deliver the necessary transparency  for ATCOs to act confidently on system recommendations~\cite{Ding2020}. Hybrid approaches combine the predictive strength of deep learning with interpretable components like attention mechanisms. These models can illustrate the input features that influenced a decision, offering a balance between accuracy and interpretability in scenarios requiring quick and reliable decision-making~\cite{Greydanus2018}.
Transparent decision-making processes also play a critical role in identifying adversarial inputs that might compromise system safety. By improving explainability, ATC systems can better detect and respond to unsafe outputs caused by adversarial attacks~\cite{Guo2021}.
Despite the advances in this field, integrating explainability into RL-based ATC systems poses several challenges. First, there is a trade-off between model complexity and interpretability. Sophisticated RL models deliver higher performance but are harder to explain, while simpler models may overlook critical aspects of ATC operations~\cite{Ghosh2021}. Real-time requirements in ATC need  both accurate and rapidly generated explanations. Finally, many ATC infrastructures rely on legacy systems, making the integration of modern RL solutions challenging without significant overhauls~\cite{Guo2022}.

\section{The simulated airspace infringement problem}
To simulate a simple airspace infringement problem, we  modeled an environment as a 2D $40 \times 40$ nautical mile grid, where an aircraft must navigate from a predefined starting position at the top center of the grid $(20, 39)$ to a target location at the bottom center $(20, 0)$. A critical feature of this environment is the inclusion of a no-fly zone (NFZ), modeled as a circular area with a center $(20,20)$ in the middle of the grid and a radius of $5$ nautical miles. The RL agent has to avoid the aircraft enters the no-fly zone and has to minimize the deviation from its original trajectory.

In  a typical scenario the agent, in each step, must take a decision about the route to the target keeping a minimum distance from the no-fly zone in red.
In each step, the state   of the problem is defined by $\mathbb{S}_k=\{P^A_k(x_k, y_k), \delta(P^A_k, P^T), \theta_k, \Phi(x_c, y_c,r)\}$, where:    
\begin{itemize}
    \item $P^A_k$ is the position of the aircraft  on the grid at the step $k$.
    \item $\delta$ represent the  Euclidean distance from the current position of the aircraft $P^A_k$ to the position of target $P^T$.
    \item $\theta_k$ is the \textit{heading angle}, that defines the direction of the aircraft, initialized randomly between 225\textdegree{} and 315\textdegree{} at the beginning of an episode.
    \item $\Phi$ represent the No-Fly Zone, defined by the coordinates of its center $(x_c,y_c)$ and its radius $r$.
\end{itemize}
At each step the agent evaluates both the progress towards the exit target and the minimal distance of its trajectory to the obstacle.

The action space $\mathbb{A}=\{\-5, 0, +5\}$ includes just 3 different actions that modify the heading angle $\theta_k$.
In particular, $-5$  turn left by 5 degrees, $0$ maintains the current heading, and $+5$ turns right by 5 degrees.
This design allows the aircraft to adjust its heading incrementally, balancing exploration and precision.

At each step the agent must take a decision about the best direction to reach the target along a trajectory that does not move too much close to the no-fly zone.
During the training phase, the agent explores the space of solutions with some random choices. For this reason, an episode can terminate with: 
\begin{itemize}
    \item \textit{Target Reached}: the aircraft exit the grid close to the target.
    \item \textit{No-Fly Zone Violation}: the aircraft enters the no-fly zone.
    \item \textit{Boundary Contact}: the aircraft touches the edge of the grid.
    \item \textit{Max Steps}: the episode exceeds 700 time steps.
\end{itemize}

\subsection{The Reinforcement Learning based solution}
For a preliminary experiment, to solve the problem of interest, we exploited a value-based reinforcement learning algorithm known as a Deep Q-Network (DQN). The DQN uses a neural network to approximate the Q-value function, denoted as $Q(s,a)$, which predicts the expected cumulative reward for taking an action $a$ in a given state $s$.
The DQN is modeled by an \textit{Input Layer} of size 6, corresponding to the parameters of the agent state; two \textit{Hidden Layers}, each with 128 neurons employing ReLU activation; an \textit{Output Layer:} of size 3, providing one Q-value per action.
The training of the DQN involves several key steps:
\begin{enumerate}
    \item \textit{Experience Replay:} Experiences $(s, a, r, s', d)$ are stored in a replay buffer, which helps to break correlations between consecutive learning samples and stabilize training.
    \item \textit{Target Network:} We maintain a copy of the Q-network, which is periodically updated to provide stable target Q-values for the loss calculation.
    \item \textit{Loss Function:} We use the Mean Squared Error (MSE) as the loss function:
    \[
    L = \mathbb{E} \left[ (Q_{\text{pred}} - Q_{\text{target}})^2 \right]
    \]
    where $Q_{\text{pred}} = Q(s, a)$ from the online network and $Q_{\text{target}} = r + \gamma \max_a Q(s', a)$ using the target network.
    \item \textit{Epsilon-Greedy Exploration:} Actions are chosen based on an $\epsilon$-greedy strategy, where $\epsilon$ is decayed over time to balance exploration and exploitation:
    \begin{itemize}
        \item With probability $\epsilon$, a random action is chosen (exploration).
        \item Otherwise, the action with the highest Q-value is selected (exploitation).
    \end{itemize}
\end{enumerate}
The learning process is based on a reward system designed to encourage the agent to reach the target while penalizing unsafe behavior.
In particular, reward function is computed as the sum of the contributions defined in Table \ref{tab:reward_contr}

\begin{table}
\centering
    \caption{Reward contribution}\label{tab:reward_contr}
    \begin{tabular}{lcl}
        \toprule
        \textit{Criteria} & \textit{Reward} & \textit{Description}\\
        \midrule
          \multirow{2}{*}{Move Reward}               &   0  & movement closer to the target\\
                                   &   -2  & movement farther from the target  \\
          \multirow{1}{*}{No-Fly Zone Penalty}              &   -200  & the aircraft touches the grid boundaries.\\
          \multirow{1}{*}{Boundary Penalty}              &   -50  &  the aircraft touches the grid boundaries\\
          \multirow{2}{*}{Target Reached}               &     $200 - (20 - x) \times 10$  & Target reached within the range $16 \leq x \leq 20$ (left side)\\ &   $100 - (15 - x) \times 10$  & Target reached within the range $11 \leq x \leq 15$ (left side)  \\ &   $30 - (10 - x) \times 5$   & Target reached within the range $6 \leq x \leq 10$ (left side)  \\ &   $5 - (5 - x)$     & Target reached within the range $1 \leq x \leq 5$ (left side)  \\ &   $200 - (x - 20) \times 10$  & Target reached within the range $21 \leq x \leq 25$ (right side)  \\ &   $100 - (x - 25) \times 10$  & Target reached within the range $26 \leq x \leq 30$ (right side)  \\ &   $30 - (x - 30) \times 5$   & Target reached within the range $31 \leq x \leq 35$ (right side)  \\ &   $5 - (x - 35)$   & Target reached within the range $36 \leq x \leq 40$ (right side)  \\&   0,40  & Target not reached or outside valid ranges  \\

        \bottomrule
    \end{tabular}
\end{table}
\subsection{The Explainability  Layer} 
\label{sect:explain}
In this simple scenario, it is easy to identify borderline results and outliers that can invalidate the tested solution, or that highlight errors or weaknesses. In more complex cases, an explainability layer allows to collect additional information about the reason of the agents decision in particular scenarios, or simply to ensure transparency in agent's decision-making.
We investigate the  utilization of a saliency map as an explainability layer.  Saliency maps highlight the importance of individual input features (e.g., position, heading angle, and distance to the target) in determining the agent's actions.
It is built computing, for each step,  the gradient of the maximum Q-value with respect to the input state:
    \[
    \text{Saliency} = \frac{\partial Q_{\text{max}}}{\partial s}
    \]
The magnitudes of the gradient indicate the sensitivity of Q-values to changes in input features. Gradients are normalized to produce interpretable saliency maps.
Saliency maps are visualized as bar charts, showing the relative importance of aircraft position (X, Y), distance to the target,  heading angle and center coordinates of the no-fly zone.
By analyzing saliency maps, we can observe how the agent prioritizes features during navigation. For example, when the aircraft is near the no-fly zone, the no-fly zone coordinates dominate the saliency map. Or, as the agent approaches the target, the distance-to-target feature becomes more prominent.
This explainability layer bridges the gap between black-box AI systems and human understanding, providing insights into how the agent balances safety (avoiding NFZ) and efficiency (moving towards the target).
\section{Experimental results}
This section evaluates the performance of the reinforcement learning (RL) agent using trajectory analysis, reward decomposition, and saliency map interpretation to provide a comprehensive understanding of its decision-making process. The results illustrate the agent’s ability to achieve its objectives while adhering to safety-critical constraints.
\begin{figure}
    \centering
    \includegraphics[width=0.6\linewidth]{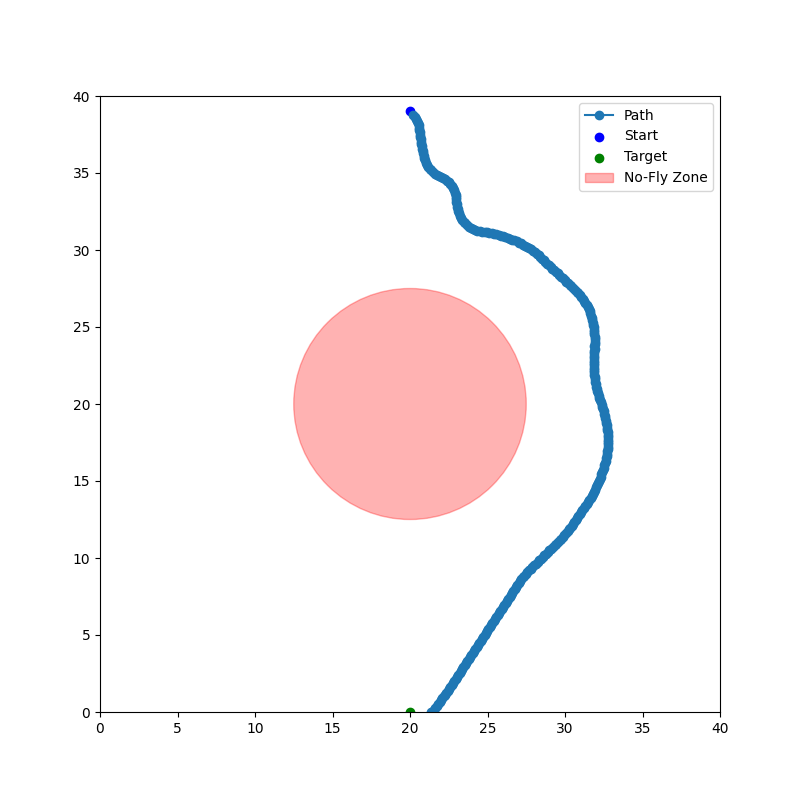}
    \caption{Episode Trajectory}
    \label{fig:episode-trajectory}
\end{figure}
The trajectory on a sample episode is shown in Figure \ref{fig:episode-trajectory}. It highlights the RL agent’s capacity to navigate a complex airspace environment effectively. The agent begins at the top of the grid and proceeds along a smooth, curved path, reaching the target located at (20,0) without entering the no-fly zone (NFZ).
The following findings highlight the capability of the RL agent to navigate a constrained airspace efficiently while adhering to safety-critical constraints:
\begin{enumerate}
\item \textit{Safety Compliance}: throughout the episode, the agent avoids  the NFZ , demonstrating its ability to navigate safely around restricted areas.
\item \textit{Goal-Oriented Navigation}: the agent successfully reaches the target, with minimal deviations or inefficiencies in its trajectory.
\item \textit{Precision and Stability}: the agent’s path is smooth and controlled, lacking abrupt or erratic adjustments, reflecting stability and effectiveness of its learned policy.
\end{enumerate}
The results of the trajectory analysis confirm that the RL agent is capable of executing complex navigation strategies in a constrained environment. 
In the following paragraphs,  we focus on the agent's decision logic discussing the results shown by the saliency map.


%

\begin{figure}
    \centering
    \includegraphics[width=0.75\linewidth]{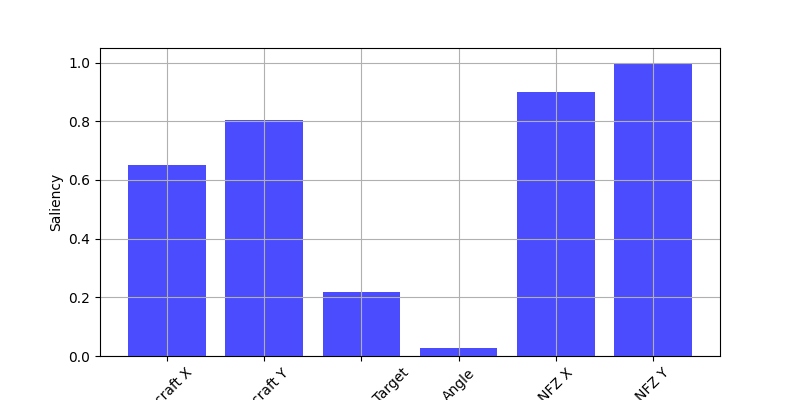}
    \caption{Saliency Map - Early Phase}
    \label{fig:saliencymap-early}
\end{figure}
\begin{figure}
    \centering
    \includegraphics[width=0.75\linewidth]{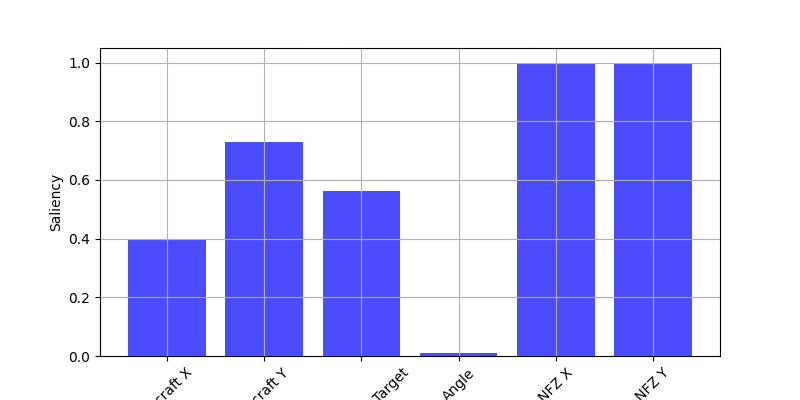}
    \caption{Saliency Map - NFZ Avoidance Phase}
    \label{fig:saliencymap-nfz-avoidance}
\end{figure}
\begin{figure}
    \centering
    \includegraphics[width=0.75\linewidth]{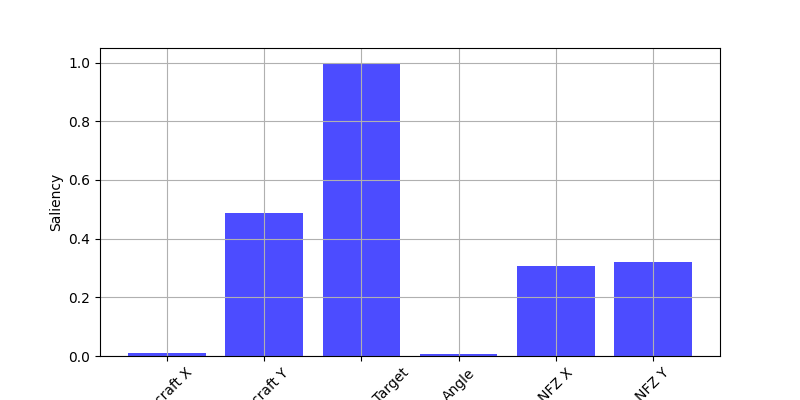}
    \caption{Saliency Map - Final Phase}
    \label{fig:saliencymap-final}
\end{figure}

Saliency maps were examined throughout the episode to provide a detailed interpretation of the agent’s decision-making process. These maps depict the temporal evolution of feature importance, offering insights into how the agent dynamically prioritizes various inputs (e.g., aircraft position, NFZ coordinates, and target location) as it navigates the environment.
The saliency map analysis reveals distinct temporal phases in the agent’s decision-making process, as illustrated from Figures \ref{fig:saliencymap-early} to Figure \ref{fig:saliencymap-final}. These figures highlight how the relative importance of various features evolves dynamically across different stages of the episode, reflecting the agent's ability to adapt its focus to meet changing environmental demands.


In the early phase (steps 0–140), the saliency map in Figure \ref{fig:saliencymap-early} reveals a more balanced prioritization between safety and goal alignment. The saliency for the NFZ coordinates remains significant, reflecting the agent’s continued focus on avoiding the restricted area as it maneuvers closer to its target. At the same time, the target feature becomes increasingly salient, indicating the agent’s dual focus on maintaining safety while advancing toward its goal. Moderated saliency for the aircraft’s X- and Y-coordinates further highlights the agent’s dynamic adjustments to refine its trajectory. These adjustments are indicative of a careful balancing act, where the agent fine-tunes its movements to simultaneously avoid obstacles and maintain efficient progress toward the target. This phase demonstrates the agent’s ability to manage competing objectives effectively, transitioning seamlessly from an initial focus on safety to a more goal-directed navigation strategy.
The critical NFZ avoidance phase (steps 150–200), shown in Figure \ref{fig:saliencymap-nfz-avoidance},  is characterized by the peak saliency observed for the NFZ coordinates (X,Y), as the agent navigates near the boundary of the restricted area. This heightened awareness underscores the importance of ensuring that the agent does not breach the NFZ, a task requiring precise control and constant spatial monitoring. As the saliency for the NFZ coordinates reaches its maximum, the agent’s decision-making is heavily influenced by safety-critical considerations during this stage. Concurrently, the saliency for the target feature begins to rise steadily, marking a transition from primarily safety-focused decisions to a more goal-oriented approach. This shift illustrates the agent’s capacity to dynamically adapt its priorities, transitioning seamlessly from obstacle avoidance to goal alignment as the immediate threat posed by the NFZ diminishes. This phase is pivotal in the episode, showcasing the agent’s ability to handle complex navigation challenges while maintaining compliance with safety constraints
In the final approach phase (steps 200–315) , the saliency map in Figure \ref{fig:saliencymap-final} shows a clear dominance of the target feature, indicating that the agent’s primary focus has shifted entirely to achieving the goal. This heightened saliency for the target reflects the agent’s precision and intent in aligning its trajectory to ensure successful completion of the task. As the agent progresses further away from the NFZ, the saliency for the NFZ coordinates diminishes significantly, confirming that the restricted area no longer poses a threat or requires active monitoring. At this stage, saliency for the aircraft’s X-coordinate becomes more prominent, supporting fine-tuned lateral adjustments that enable the agent to align accurately with the target position. This phase underscores the agent’s ability to execute precise, goal-oriented behaviors in the final steps of the trajectory, culminating in a successful episode. The saliency patterns during this phase reflect the agent’s confidence in its trajectory and its capability to complete the task without additional safety concerns.

\section{Conclusion}
\label{sect:coclusion}
The integration of trajectory analysis, reward decomposition, and saliency maps offers a multi-faceted evaluation of the RL agent’s performance, with a primary focus on enhancing explainability. In safety-critical applications, such as air traffic management, the ability to interpret and understand the decision-making processes of RL systems is paramount for ensuring trust and accountability. This study leverages these analytical tools to provide transparency into the agent’s dynamic prioritization of features and its adherence to safety constraints. By examining the agent’s trajectory, the sources of reward contributions, and the evolution of saliency over time, we demonstrate how explainability can bridge the gap between the complexity of RL models and the operational demands of real-world environments.
\begin{acknowledgement} This work has been partially supported by the REDRAW  research project (P2022MWE3S - Prin 2022 PNRR, DR n. 1409 of 14-09-2022) funded by the Italian Ministry of Research and by the European Union, and by project SERICS (PE00000014) under the MUR National Recovery and Resilience Plan funded by the European Union - NextGenerationEU, and by JARVIS, a SESAR Industrial Research project within the Digital European Sky programme - grant ID 101114692 (https://www.sesarju.eu/projects/JARVIS) 
\end{acknowledgement}
\bibliographystyle{spmpsci}
\bibliography{chapter.bib}
\end{document}